\documentclass[letterpaper,10pt]{article}
\usepackage[dvips]{graphicx}
\usepackage[dvips]{floatflt}
\usepackage{theorem}

\newcommand{\be}{\begin{equation}}
\newcommand{\ee}{\end{equation}}
\newcommand{\beq}{\begin{equation}}
\newcommand{\eeq}{\end{equation}}
\newcommand{\bed}{\begin{displaymath}}
\newcommand{\eed}{\end{displaymath}}
\newcommand{\beqa}{\begin{eqnarray}}
\newcommand{\eeqa}{\end{eqnarray}}
\newcommand{\beqann}{\begin{eqnarray*}}
\newcommand{\eeqann}{\end{eqnarray*}}
\newcommand{\bseq}{\begin{subequation}}
\newcommand{\eseq}{\end{subequation}}

\newcommand{\ba}{\begin{array}}
\newcommand{\ea}{\end{array}}

\newcommand{\negr}[1]{{\bf {#1}}}

\makeatletter
\theoremstyle{plain}

\newtheorem{Def}{Definition}

\def\@normalsize{\@setsize\normalsize{10pt}\xpt\@xpt
\abovedisplayskip 3pt plus2pt minus2pt
\belowdisplayskip\abovedisplayskip
\abovedisplayshortskip \z@ plus3pt
\belowdisplayshortskip 5pt plus2pt minus5pt
\let\@listi\@listI}

\def\figurename{\small Figure}
\def\subsize{\@setsize\subsize{10pt}\xipt\@xipt}
\def\section{\@startsection {section}{1}{\z@}{19pt plus 1pt minus 5pt}
{9pt plus 1pt minus 5pt}{\bf}}
\def\subsection{\@startsection {subsection}{2}{\z@}{9pt plus 1pt minus 4pt}
{8pt plus 1pt minus 4pt}{}}
\def\subsubsection{\@startsection {subsubsection}{2}{\z@}{8pt plus 1pt minus 4pt}
{2pt plus 1pt minus 1pt}{\em}}
\makeatother

\begin{document}
\date{}  
\title {\noindent\bf WORKSPACE AND ASSEMBLY MODES IN FULLY-
\goodbreak\noindent PARALLEL MANIPULATORS: A DESCRIPTIVE STUDY}
\author{\begin{tabular}[t]{ c l}
~~~~~~  &  { Ph. WENGER AND D. CHABLAT} \\
~~~~~~  &  {\em Institut de Recherche en Cybern\'etique de Nantes} \\
~~~~~~  &  {\em 1, rue de la No\"e, 44321 Nantes, France} \\
~~~~~~  &  {\bf email: Philippe.Wenger@lan.ec-nantes.fr} \\
\end{tabular}}
\maketitle
{\noindent\bf Abstract~:} {The goal of this paper is to explain, using a typical example, the distribution of the different assembly modes in the workspace and their effective role in the execution of trajectories. The singular and non-singular changes of assembly mode are described and compared to each other. The non-singular change of assembly mode is more deeply analysed and discussed in the context of trajectory planning. In particular, it is shown that, according to the location of the initial and final configurations with respect to the uniqueness domains in the workspace, there are three different cases to consider before planning a linking trajectory.}
\section{Introduction}
\noindent Most of the active research work carried out word wide in the field
of parallel manipulators have focused on a particularly challenging
problem, namely, solving the forward kinematic problem, or, in
other words, finding the different poses of the mobile
platform (the assembly modes) in function of the positions of the
actuated joints. A second interesting problem has been the
evaluation and optimization of the workspace of parallel
manipulators (Merlet, 97) and (Gosselin, 88). It is worth noting
that the forward kinematic problem and the workspace analysis are
most often treated separately, although they are closely linked to
each other. It is well known that parallel manipulators have
singularities in their workspace where stiffness is lost (Gosselin,
90). These singularities coincide with the set of configurations in
the workspace where two direct kinematic solutions meet (Hunt, 93).
On the other hand, it was recently shown that the change of
assembly mode could also be accomplished without passing through a
singularity (Innocenti, 92). This result gave rise five years later
to a theoretical work with the concepts of characteristic surfaces
and uniqueness domains in the workspace (Wenger, 97). However, no
result has been provided as to the practical interest of these
concepts, especially in trajectory planning purposes. The goal of
this paper is to explain, using a typical example, the distribution
of the different assembly modes in the workspace and their
effective role in the execution of trajectories. The singular and
non-singular changes of assembly mode are described and compared to
each other. The non-singular change of assembly mode is more deeply
analysed and discussed in the context of trajectory planning. This
paper is organized as follows. Section 2 recalls the necessary
definitions. Section 3 develops the descriptive analysis of this
work. A 3-DOF planar manipulator is used as illustrative example
all along this study. The 3-D workspace is evaluated and depicted
using octree structures. Schematic diagrams are additionally used
to improve the legibility of the explanations. Section 4 yields
some ideas on how to use the results of section 3 and the
uniqueness domains in the purpose of trajectory planning. Section 5
concludes this paper.
\section{Preliminaries}
In this paragraph, some definitions permitting to understand this
paper are quoted.
\subsection{FULLY PARALLEL MANIPULATORS}
\begin{Def}
A fully parallel manipulator is a mechanism that includes as many
elementary kinematic chains as the mobile platform does admit
degrees of freedom. Moreover, every elementary kinematic chain
possesses only one actuated joint. Besides, no segment of an
elementary kinematic chain can be linked to more than two bodies
(Merlet, 97).
\label{Definition:Fully_Parallel_Manipulator}
\end{Def}
In this study, kinematic chains, or legs (Angeles, 97), are always
independent.
\subsection{KINEMATIC RELATIONS AND SINGULARITIES}
For a manipulator, the relation permitting the connection of input
values (\negr q) with output values (\negr X) is the following
\be
        F(\negr X, \negr q)=0
        \protect\label{equation:the_kinematic}
\ee
\negr q is the vector of actuated joints and \negr X is the vector of
configurations of the output link (mobile platform). The set of all
admissible \negr q will be referred to as the {\em joint space} and
the set of all reachable \negr X is the {\em workspace}.
Differentiating equation (\ref{equation:the_kinematic}) with
respect to time leads to the velocity model
\be
     \negr A \negr t + \negr B \dot{\negr q} = 0
\ee
Moreover, \negr A  and \negr B are respectively the
direct-kinematics and the inverse-kinematics matrices of the
manipulator. A singularity occurs whenever \negr A or
\negr B, (or both) that can no longer be inverted. Three
types of singularities exist in general (Gosselin, 90): $det(\negr
A)= 0$ or $det(\negr B)= 0$ or $det(\negr A)= 0$ and $det(\negr B)=
0$.
\par
In this study, only the singularities for which $det(\negr A)= 0$
(referred to as parallel singularities (Wenger, 97)), will be of
interest. The corresponding singular configurations are located
inside the workspace. They are particularly undesirable because the
manipulator can not resist any effort.
\subsection{NOTION OF ASPECT FOR FULLY PARALLEL MANIPULATORS}
The notion of aspect was introduced by (Borrel,86) to cope with the
existence of multiple inverse kinematic solutions in serial
manipulators. In this paper, we will use the notion of aspect
defined in (Wenger, 97) for parallel manipulators with only one
inverse kinematic solution, which are the object of this study.
\par
\begin{Def}
\label{definition:Aspect}
The {\em aspects} $\negr {WA}_i$ are defined as the maximal sets
such that
\end{Def}
\begin{itemize}
\item $\negr {WA}_i \subset W$;
\item $\negr {WA}_i$ is connected;
\item $\forall \negr X \in \negr {WA}_i,  Det(\negr A) \neq 0$
\end{itemize}
In other words, the  aspects $\negr {WA}_i$ are the maximal
singularity-free regions in the workspace.
\section{Descriptive Analysis}
\subsection{MANIPULATOR EXAMPLE USED}
\begin{floatingfigure}[r]{50mm}
\begin{center}
     \includegraphics[width= 50mm,height= 40mm]{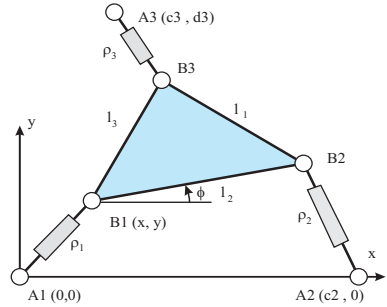}
     \caption{A 3 - RPR parallel manipulator}
     \protect\label{figure:manipulateur_RPR}
\end{center}
\end{floatingfigure}
The $3-RPR$ parallel manipulator shown in figure
\ref{figure:manipulateur_RPR} has been frequently studied (see for
instance (Gosselin, 88) and (Innocenti, 92)). The input joint
variables are the three prismatic joints. The output variables are
the position and the orientation of the platform in the plane. The
passive joints will always be assumed unlimited in this study. The
limits of the prismatic actuated joints are those chosen in
(Innocenti, 92) and (Wenger, 97) ($10.0 \leq
\rho_i
\leq 32.0$). The dimensions of the platform are the same as in
(Merlet, 97) and in (Innocenti, 92):
\begin{itemize}
\item $A_1= (~0.0~;~0.0)$ \quad $B_1B_2= 17.04$;
\item $A_2= (15.91;~0.0)$ \quad $B_1B_2= 16.54$;
\item $A_3= (~0.0~;10.0)$ \quad $B_1B_2= 20.84$;
\end{itemize}
\subsection{WORKSPACE, SINGULARITIES AND ASPECTS}
\begin{figure}[hbt]
    \begin{center}
    \begin{tabular}{cccc}
       \begin{minipage}[t]{60 mm}
           \centerline{\hbox{\includegraphics[width= 49mm,height= 51mm]{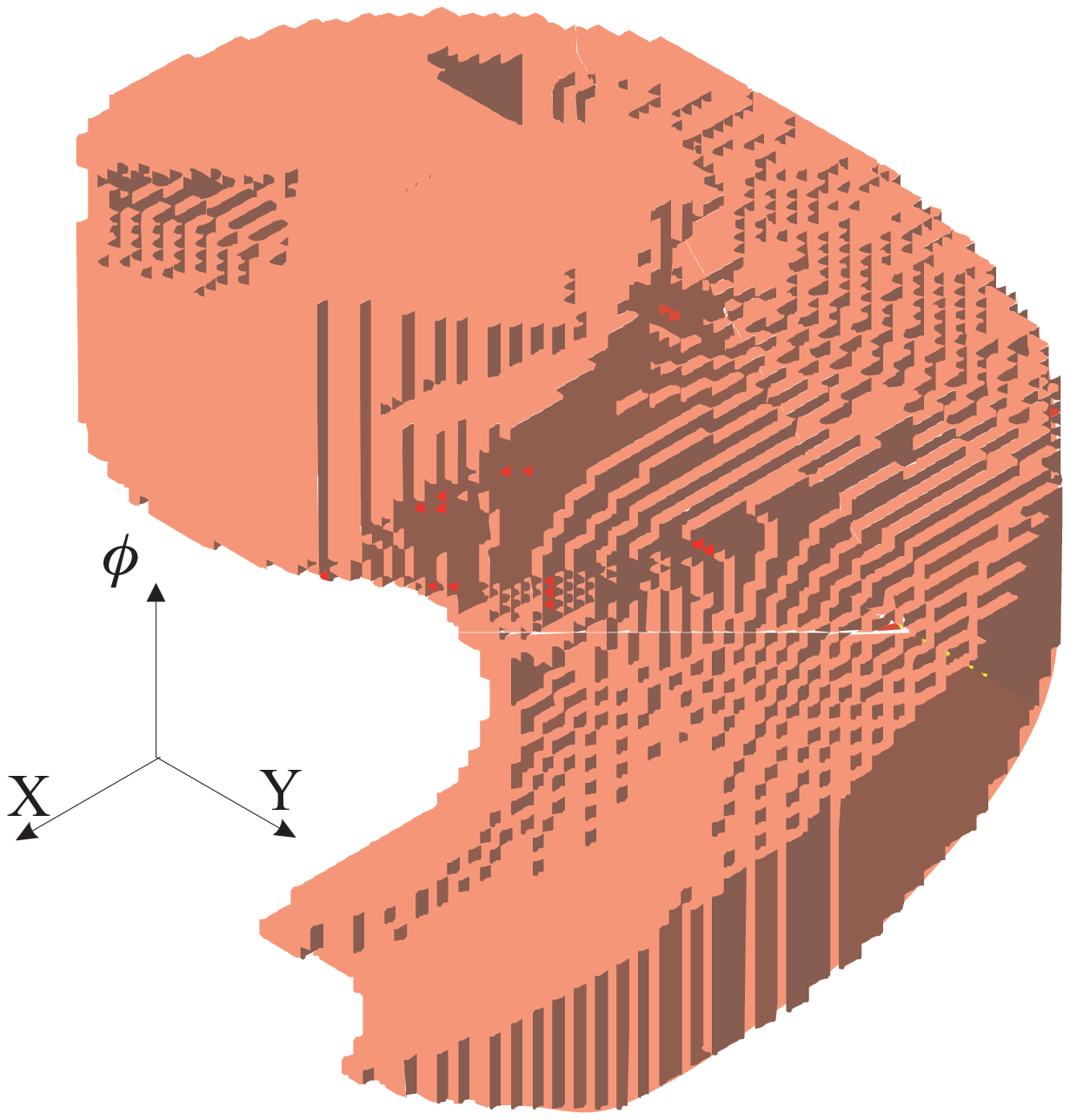}}}
           \caption{Workspace}
           \protect\label{figure:workspace}
       \end{minipage} &
       \begin{minipage}[t]{60 mm}
           \centerline{\hbox{\includegraphics[width= 48mm,height= 47mm]{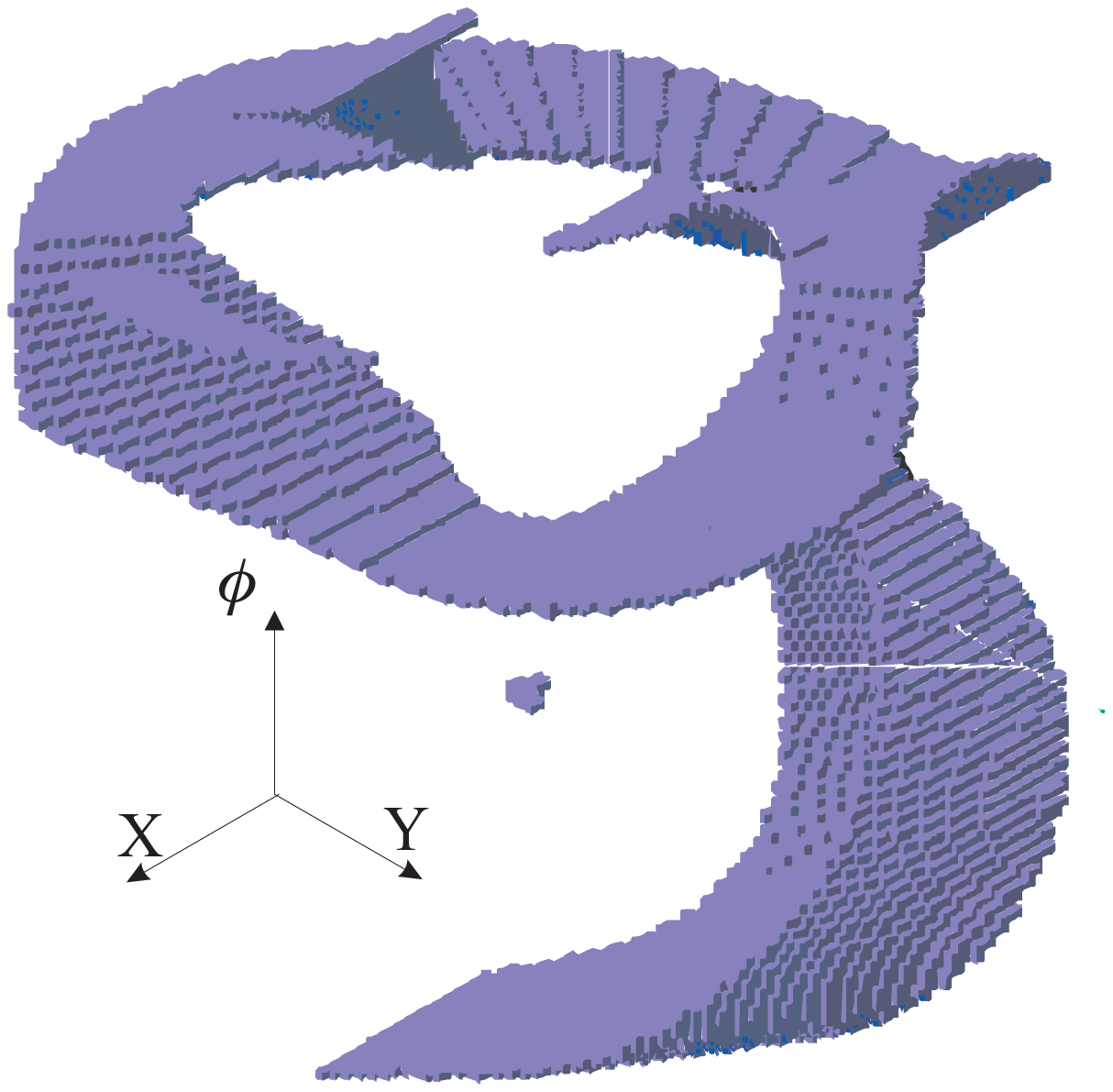}}}
           \caption{The singularity surface}
           \protect\label{figure:singularity}
       \end{minipage}
    \end{tabular}
    \end{center}
\end{figure}

\begin{figure}[htp]
    \begin{center}
    \begin{tabular}{cccc}
       \begin{minipage}[t]{60 mm}
           \centerline{\hbox{\includegraphics[width= 48mm,height= 50mm]{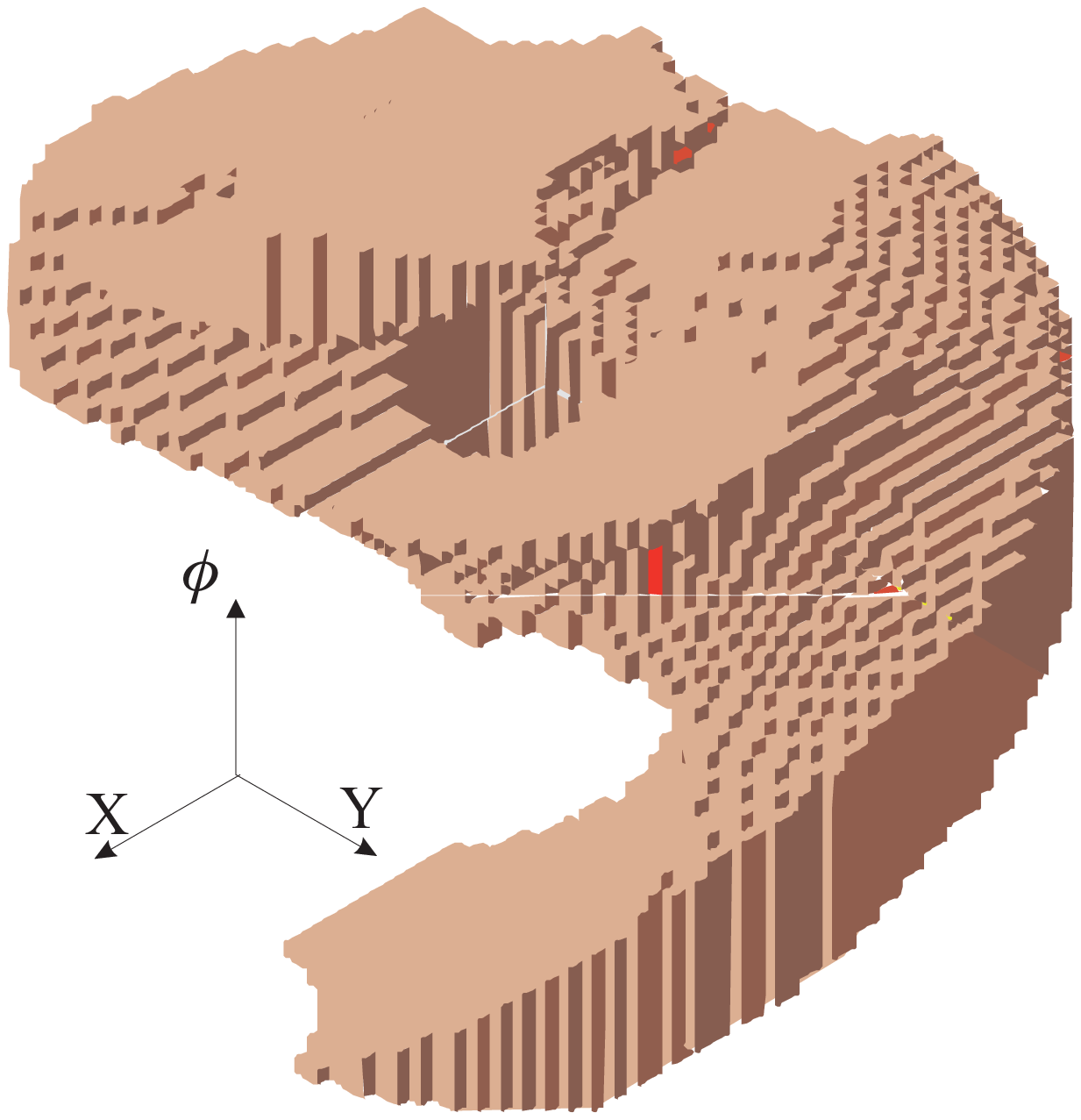}}}
           \caption{First aspect in the Workspace ${\bf WA_1}$}
           \protect\label{figure:aspect1}
       \end{minipage} &
       \begin{minipage}[t]{60 mm}
           \centerline{\hbox{\includegraphics[width= 45mm,height= 50mm]{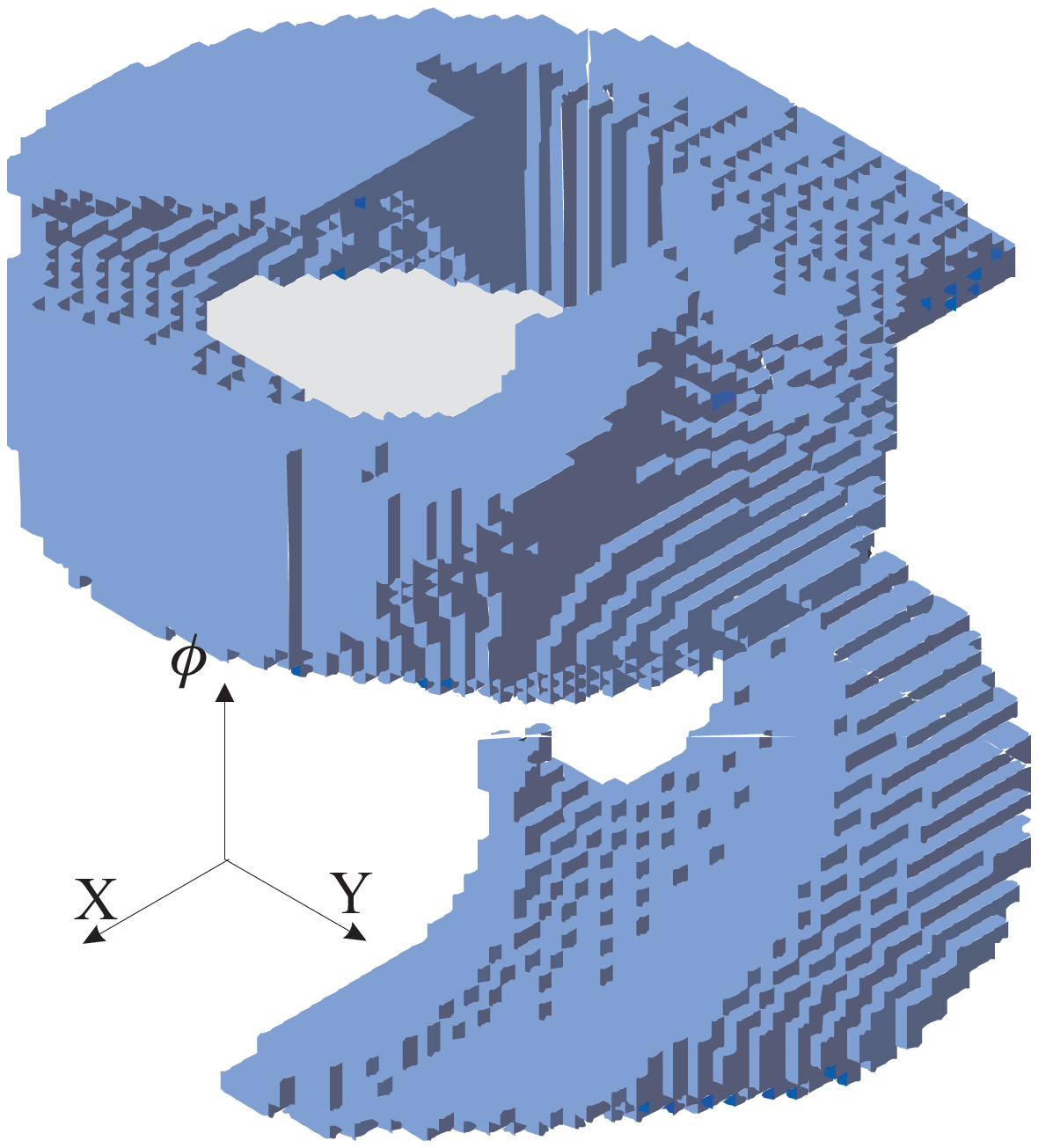}}}
           \caption{Second aspect in the Workspace ${\bf WA_2}$}
           \protect\label{figure:aspect2}
      \end{minipage}
    \end{tabular}
    \end{center}
\end{figure}

The workspace of the manipulator at hand is 3-dimensional. A 3-D
representation can be made in which the vertical axis represents
the orientation of the output link in the (x,y)-plane. The
workspace is usually analysed in position only (see (Merlet, 98)
for instance) and without consideration of the assembly modes.
Figure \ref{figure:workspace} shows the full 3-D workspace modelled
with octree structures. Octrees have nice interesting properties
like the existence of an implicit adjacency graph which enables
simple connectivity analyses (Wenger, 97) (Samet,79). An efficient
calculation technique using octrees has been developed by the
authors (Chablat, 97) but is not reported here for lack of space.
The dimensions and the displacement ranges of the linear actuators
are such that the dextrous workspace is non zero. That is, the
output link can admit any orientation. In other words, the upper
and lower sides of the workspace actually coincide (the workspace
has the structure of a torus). Another important feature is that a
singular surface lies inside this workspace and divides it into two
adjacent aspects (Sefrioui, 92). When the manipulator output link
lies on this singular surface, the manipulator is in a
configuration such that the axes on the linear actuators intersect
at a common point.
Figure \ref{figure:singularity} shows the singular surface and the
two aspects are depicted in figure \ref{figure:aspect1} and
\ref{figure:aspect2}, respectively. It is well known that our
manipulator admits up to 6 direct kinematic solutions (Gosselin,
92). It was shown in (Wenger, 97) that there are 3 solutions in
each aspect. In other words, up to 3 assembly modes are available
in a same singularity-free region of the workspace.
\subsection{\uppercase{Changing assembly mode and the structure of the \goodbreak workspace}}
An assembly mode is associated with a solution to the forward
kinematics. Changing assembly mode means going from one solution to
another. In practice, a change of assembly mode may occur during
the execution of a trajectory between two configurations in the
workspace which are not necessarily associated with the same input.
\par
Both singular and non-singular assembly mode changes are possible
for the manipulator studied. We begin the analysis with the
non-singular change of assembly mode. Then, since it was shown
recently that it was possible for a parallel manipulator to make it
go through a singularity (Nenchev, 97), singular change of assembly
mode will be also discussed in this section.
\subsubsection{Non-singular change of assembly mode}
\begin{figure}[hbt]
\begin{center}
    \begin{tabular}{cccc}
       \begin{minipage}[t]{60 mm}
           \centerline{\hbox{\includegraphics[width= 62mm,height= 49mm]{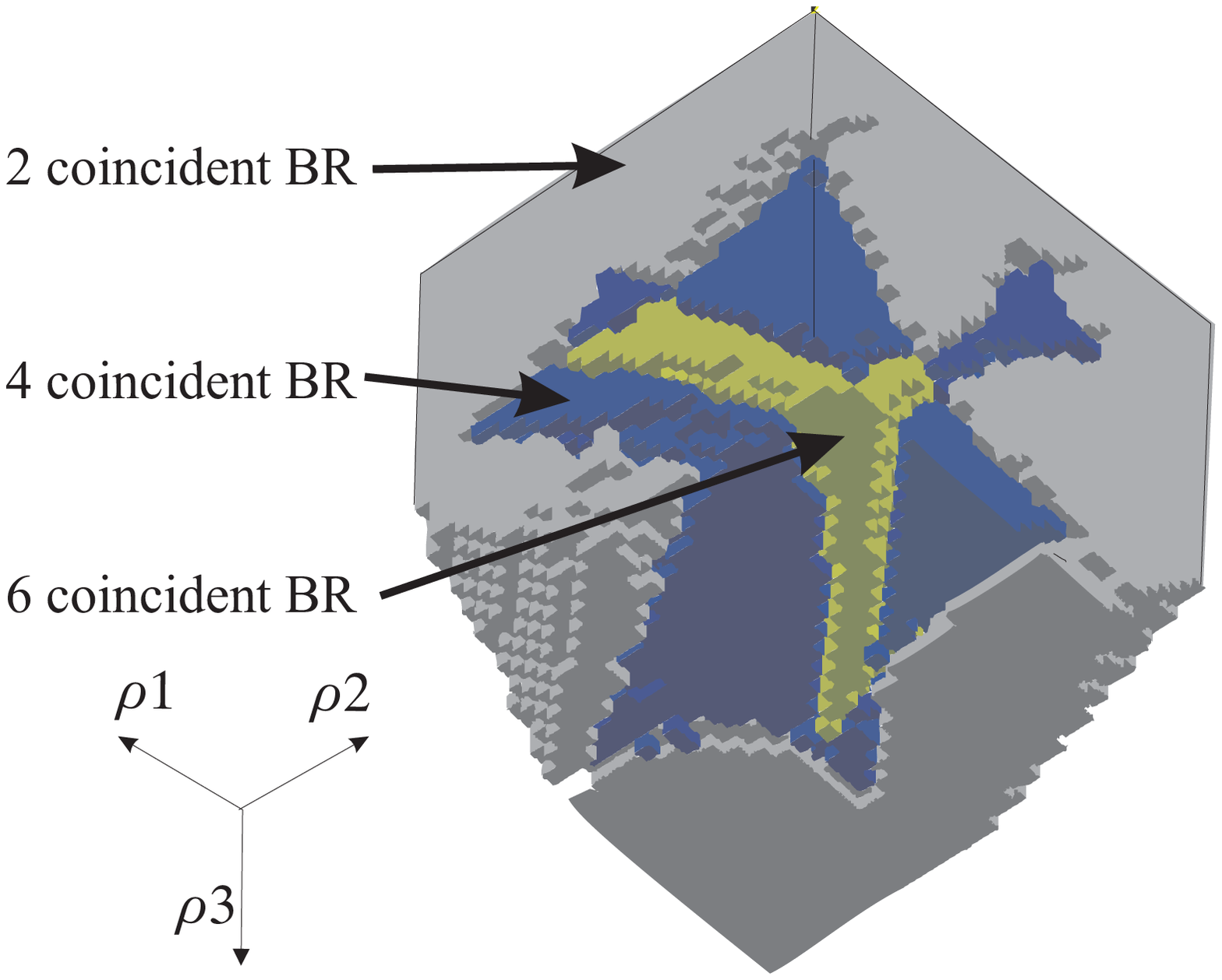}}}
           \caption{The basic regions in joint space (BR)}
           \protect\label{figure:joint_space}
       \end{minipage} &
       \begin{minipage}[t]{60 mm}
           \centerline{\hbox{\includegraphics[width= 61mm,height= 42mm]{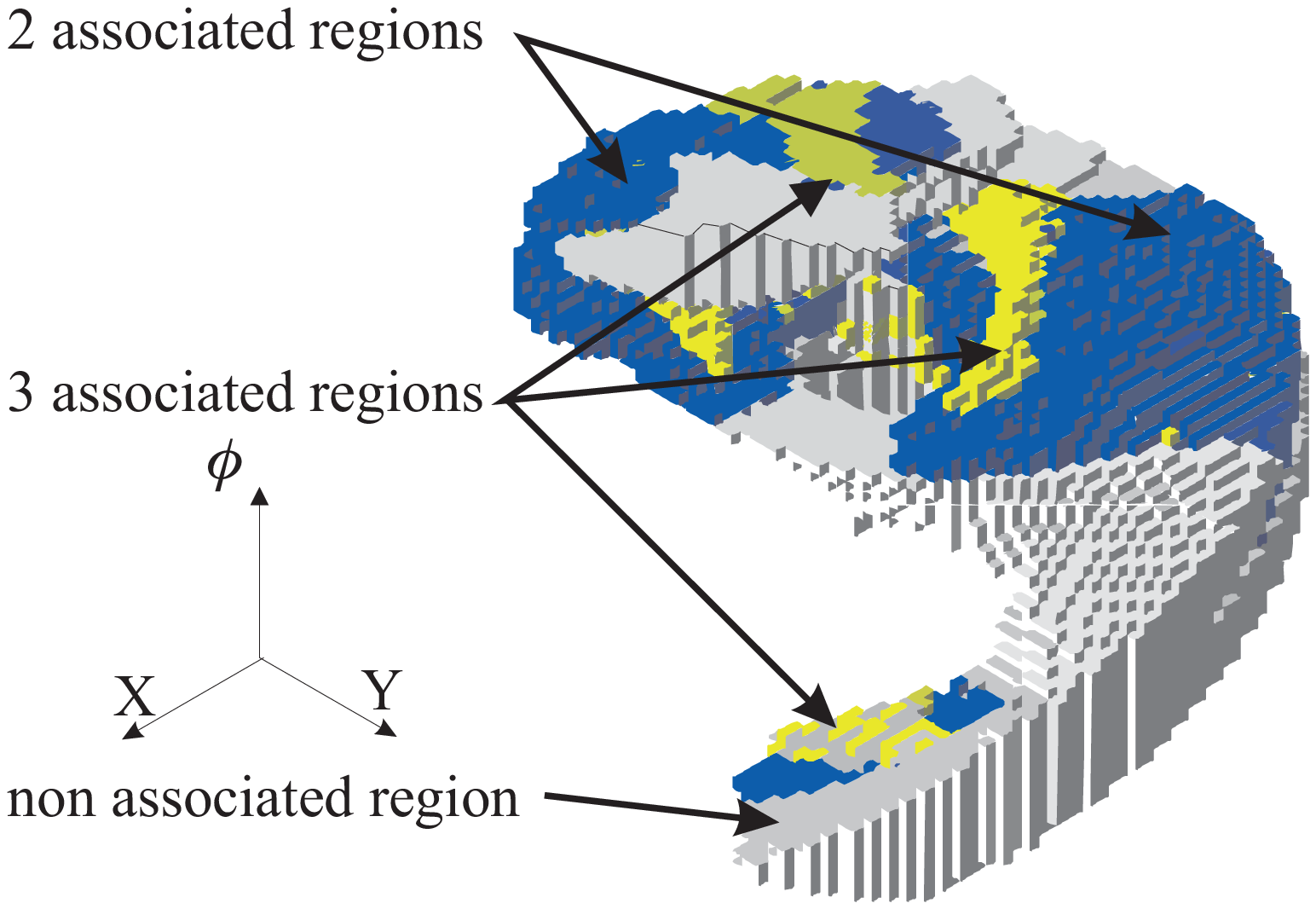}}}
           \caption{The corresponding basic regions in one aspect of the workspace}
           \protect\label{figure:work_space}
      \end{minipage}
    \end{tabular}
\end{center}
\end{figure}

For the purpose of trajectory planning, it is interesting to
investigate more deeply when and how a parallel manipulator changes
assembly mode without crossing singularities. To begin with, let us
examine the topological structure of the workspace. The
singularities divide the workspace into aspects and the
characteristic surfaces induce a partition of each aspect into a
set of regions (the basic regions, see (Wenger, 97)). On the other
hand, the singularities also divide the joint space into several
regions. Each region of the joint space is characterized by a
number, say p, of direct kinematic solutions (or assembly modes)
and can be interpreted as being composed of a stack of p coincident
basic components. These coincident p basic components are separated
under the action of the direct kinematics to form p disjoint and
non adjacent basic regions in the workspace. Such basic regions
will be called {\em associated regions} and are physically
associated to the different admissible assembly modes in one
aspect. In figure \ref{figure:joint_space}, the joint space is
composed of one region with 6 coincident basic components, four
regions with 4 coincident basic components and one region with 2
coincident basic components.
\par
\begin{figure}[hbt]
  \begin{center}
     \includegraphics[width= 110mm,height= 55mm]{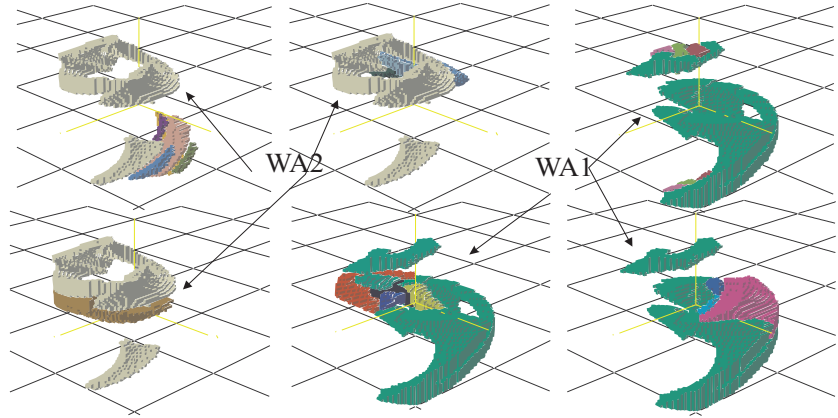}
     \caption{The uniqueness domains in the workspace}
     \protect\label{figure:trav}
  \end{center}
\end{figure}

In the workspace, the basic components are separated and equally
distributed in the two aspects to form, in each aspect, one set of
3 associated regions, one set of 2 associated regions and one basic
region which is not associated with another region in the same
aspect (figure \ref{figure:work_space}). The uniqueness domains are
composed of regions which are not associated. By definition, there
is a one-to-one correspondence between one uniqueness domain and
the joint space. Figure \ref{figure:trav} shows the different
uniqueness domains obtained for the manipulator studied: there are,
in this case, 3 uniqueness domains in each aspect (${\bf WA_1}$ and
${\bf WA_2}$). The important property of a uniqueness domain is
that {\em any displacement of the output link can be accomplished
in a whole uniqueness domain while never changing assembly mode}.
Since the uniqueness domains are, by definition, the maximal sets
associated with one assembly mode, it can be claimed that,
consequently, the {\em uniqueness domains are the maximal regions
of the workspace where all the displacements of the output link can
be accomplished while never changing assembly mode.}
\par
In contrast, the associated regions are those regions which cannot
be linked without changing assembly mode. This means that if the
output link must move between two configurations lying in two
associated regions, the only way to perform this task is to execute
a non-singular assembly mode changing trajectory. Is it always
possible to move between two associated regions? The answer is yes
if and only if the two regions belong to a same path-connected
component of the workspace. As a matter of fact, it is clear that
for a motion to exist between two configurations in the workspace,
these configurations should belong to a same connected component.
Conversely, if it is so, there is a continuous path in the
workspace which maps uniquely onto a continuous path in the joint
space because the inverse kinematics is continuous and admits only
one solution.
\par
We can set the following important result: {\em Each
singularity-free domain (or aspect) of the workspace is composed of
several uniqueness domains which are associated with one unique
assembly mode. Given two points in an aspect, it is always possible
to move between these points if they belong to a same connected
component. These points can always be linked without changing
assembly mode if they belong to a same uniqueness domain. In
contrast, a non-singular assembly mode trajectory will have to be
executed if the two points belong to the associated regions of two
distinct uniqueness domains.}
\par
\begin{figure}[hbt]
     \begin{center}
        \includegraphics[width= 79mm,height= 35mm]{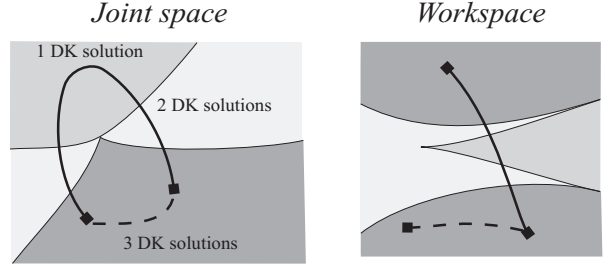}
        \caption{Right (solid) and wrong (dotted) trajectories (scheme)}
        \protect\label{figure:trajectory_1}
     \end{center}
\end{figure}

This result is of primary importance since according to the
location of configurations in the workspace, the trajectory
planning will not have to be equally treated. In the case of two
configurations lying in a same uniqueness domain, any continuous
trajectory is feasible in this domain since, by definition, there
is a one-to-one correspondence between the uniqueness domain and
the joint space. On the other hand, if the two points belong to two
distinct uniqueness domains (that is, in two associated regions),
the problem is different since the trajectory will have to enable a
change of assembly mode. Now, we will explain how a non-singular
change of assembly mode can be realised.  When a motion is
prescribed between two associated regions, a specific trajectory
must be executed in the joint space. This trajectory must link the
two coincident basic components of the joint space corresponding to
the two associated regions in the workspace. It can be shown, using
a similar analysis as for cuspidal serial manipulators (Wenger,
96), that the coincident basic components are not directly
connected by their boundaries, but through an adjacent component
where the number of direct kinematic solutions is lower. Thus, a
typical non-singular assembly mode changing trajectory is not a
straightforward path inside the joint space. Instead, it will be a
trajectory that will leave the initial basic component through a
boundary surface, transits through an intermediate adjacent
component and finally enters the goal basic component (which is
actually coincident to the initial one) by crossing a different
boundary surface. In the workspace, a non-singular assembly mode
changing trajectory always crosses at least two characteristic
surfaces (which are associated with the boundary surfaces crossed
in the joint space) and a basic region which is not associated with
the initial and final regions. Such a manoeuver is illustrated in
solid lines in figure \ref{figure:trajectory_1}. In the same
figure, the dotted lines show the path in the workspace that would
result from a direct trajectory in the joint space. This path
starts from the initial prescribed configuration but does not reach
the desired goal configuration since the assembly mode has not
changed. From a practical point of view, this means that
trajectories should be planned in the workspace rather than in the
joint space. We will come back on this point in section 3.4.
\subsubsection{Singular change of assembly mode}
\begin{figure}[hbt]
    \begin{center}
    \begin{tabular}{cccc}
       \begin{minipage}[t]{60 mm}
           \centerline{\hbox{\includegraphics[width= 49mm,height= 29mm]{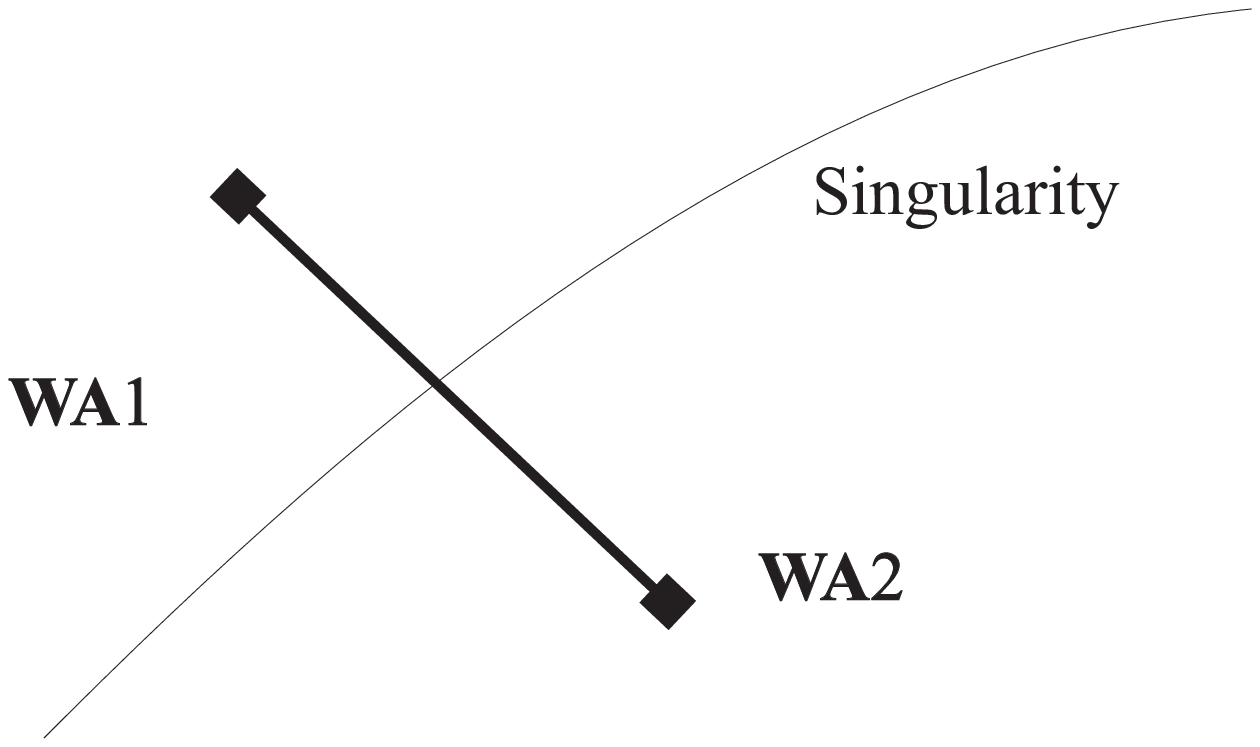}}}
           \caption{Two configurations in two distinct aspects in workspace}
           \protect\label{figure:2_configurations}
       \end{minipage} &
       \begin{minipage}[t]{60 mm}
           \centerline{\hbox{\includegraphics[width= 38mm,height= 28mm]{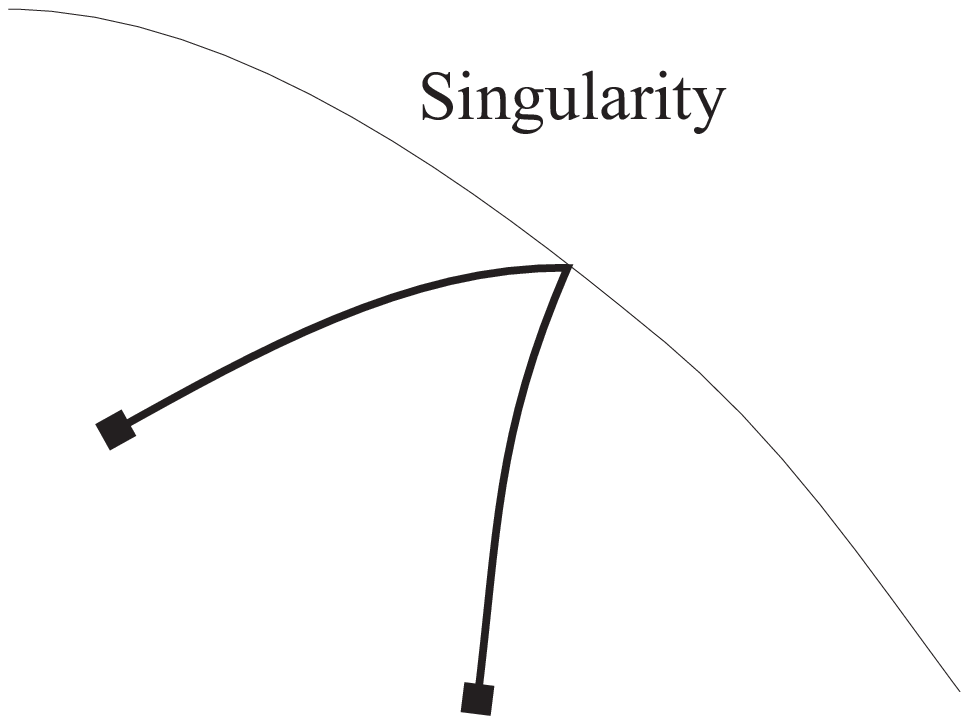}}}
           \caption{Singular assembly mode changing joint trajectory}
           \protect\label{figure:trajectory_joint_space}
       \end{minipage}
    \end{tabular}
    \end{center}
\end{figure}

A singular change of assembly mode implies going through a
singularity where the manipulator looses stiffness. For a long
time, it has been commonly argued that such a manoeuver was not
possible in practice because of the lost of control at the
singularity. However, Nenchev et al have shown recently that it was
possible to control the motion of a parallel manipulator through a
singularity under certain conditions on the instantaneous direction
of motion of the output link (Nenchev, 97). A singular change of
assembly mode will be necessary when the two configurations to be
linked are located in two distinct aspects in the workspace, like
in figure \ref{figure:2_configurations}. When the output link
crosses a singularity, it can be shown that the corresponding joint
trajectory ``reflects back'' against a boundary in the joint space,
which is the image of the singularity crossed in the workspace
(figure \ref{figure:trajectory_joint_space}). As a matter of fact,
given a vector of actuated joint position near a singularity,
(Innocenti, 92) has shown that there are two ``related
configurations'' in the workspace which are symmetrically located
with respect to the singularity in the workspace and these two
related configurations merge at the singularity. Thus, in contrast
with 3.3.1, a singular assembly mode changing trajectory gives rise
to a joint trajectory which links two coincident basic components
directly by their boundary. This joint trajectory is analogous to
the cartesian singular change of posture trajectory of a ``Puma''
serial manipulator when it switches from ``elbow up'' to ``elbow
down''.
\subsubsection{Application to the trajectory planning}
This section only provides some key ideas on how the previous
analysis and the octree model of spaces can be applied to the
trajectory planning. We assume here for more simplicity that there
is no collision so that the aspects and the uniqueness domains are
connected (Wenger, 97). However, we know that the workspace is not
necessarily connected even if all motions are free of collisions.
So, the first thing to do is to verify whether the two
configurations belong to the same connected component of the
workspace. This test can be done easily with the octree model of
the workspace using a path-connectivity analysis algorithm (Samet,
79). If the two configurations are not in a same connected
component, no trajectory can be found. If yes, according to the
preceding analysis, there are three main cases to consider:
\begin{itemize}
\item 1. the two configurations are in two distinct aspects
\item 2. the two configurations are in a same aspect but in two distinct uniqueness domains
\item 3. the two configurations are in a same uniqueness domain.
\end{itemize}
All cases can be easily checked using the octree models of the
aspects and of the uniqueness domains. In the first case, we are in
a situation where a singular change of assembly mode is necessary.
In this case, the analysis of (Nenchev, 97) should be used to build
a path which permits a feasible control law. In the second case, a
non-singular change of assembly mode will have to be executed. A
path must be constructed in the workspace. For the calculation of
the corresponding path in the joint space, a simple method is to
compute the inverse kinematics for a series of configurations by
discretisation of the workspace path, without omitting those
configurations where the output link crosses the characteristic
surfaces. The last case is the nicest situation since we can remain
in a same uniqueness domain and the kinematics is one-to-one
between the joint space and the uniqueness domain. In this case, it
is possible to compute a feasible trajectory directly in the joint
space, which is more convenient for optimizing certain criteria
(like the actuator torques or the cycle time for instance). Under
all circumstances, the search for a feasible path can be achieved
with classical tools using the octree structures (see (Faverjon,
84) for instance). The well-known A* algorithm can be used,
together with a path cost estimation procedure (on the basis of a
chosen relevant criteria).
\section{Conclusion}
The descriptive study provided in this paper has shown the interest
of defining and calculating the aspects and the uniqueness domains
in the workspace for trajectory planning. A change of assembly mode
will be necessary when and only when the initial and goal
configurations are not in a same uniqueness domain. Such manoeuvres
are possible if and only if the two configurations are in a same
connected component of the workspace. A non-singular change of
assembly mode will occur between two configurations which are in
two distinct uniqueness domains but in a same aspect while a
singular change of assembly mode is only necessary when the
configurations are in two distinct aspects. It is of interest to
know in advance whether two given configurations will be linked
with or without a change of assembly mode. It is more desirable to
plan trajectories that keep the same assembly mode since such
trajectories will generally lead to smoother displacements of the
actuated joints. That is, the working configurations should be
located in a same uniqueness domains, which could be garanteed by a
proper placement and/or design of the manipulator. This work brings
also some new ideas for the optimal design of parallel
manipulators: it is more convenient to have a manipulator with
large uniqueness domains rather than with a large workspace since a
large workspace can be composed of many small uniqueness domains. A
planar manipulator has been used to illustrate this work but the
tools developed with octree structures permit to treat any 3-DOF
fully-parallel manipulators like spatial positioning manipulators.
Work is under development to take into account the effects of
collisions (between legs and with obstacles) in the calculation of
the uniqueness domains.
\section{References}
{\small Angeles, J., (1997),
 {\em Fundamentals of Robotic Mechanical Systems}, SPRINGER.
\par
Borrel, P., (1986), A study of manipulator inverse kinematic
solutions with application to trajectory planning and workspace
determination, Proceeding IEEE International Conference on Robotic
And Automation,
 pp 1180-1185.
\par
Chablat, D., and Wenger, Ph., (1997) Domaines d'unicit\'e des
manipulateurs parall\`eles, IRCyN Internal Report,
$N^{\circ}96.13$, Nantes.
\par
Faverjon, B., (1998), Obstacle avoidance using an octree in the
configuration space of a manipulator, 1st I.E.E. Conference on
Robotics and Automation, Atlanta, pp.~504-512.
\par
Gosselin, C., and Angeles, J., (1988), The Optimum Kinematic Design
of a Planar Three-Degree-of-Freedom Parallel Manipulator, ASME,
Journal of Mechanisms, Transmissions, and Automation in Design,
Vol. 110, March.
\par
Gosselin, C., and Angeles, J., (1990), Singularity analysis of
closed-loop kinematic chains, IEEE Transactions On Robotics And
Automation, Vol.~6, No.~3.
\par
Gosselin, C., Sefroui, J., and Richard M. J., (1992), Solutions
polynomiales au probl\`eme de la cin\'ematique directe des
manipulateurs plans \`a trois degr\'es de libert\'e, Mechanical and
Machine Therory, Vol. 27, No. 2, pp 107-119.
\par
Hunt, K. H., and Primrose, E. J. F., (1993), Mechanical and Machine
Theory, Vol. 28, No 1, pp. 31-42.
\par Innocenti C., and
Parenti-Castelli V., (1992), Singularity-free evolution from one
configuration to another in serial and fully-parallel manipulators,
Proceeding ASME Design Technical Conferences, DE-Vol. 45, Robotics,
Spatial Mechanisms and Mechanical Systems, pp.~553-560, ASME.
\par
Merlet, J-P., (1997), {\em Les robots parall\`eles}, HERMES,
seconde \'edition, Paris.
\par
Merlet, J-P., Gosselin, C., and Mouly, N., (1998), Workspaces of
Planar Parallel Manipulators, Mechanical and Machine Theory, Vol.
33, No. 1/2, pp 7-20.
\par
Nenchev, D.N., Bhattacharya, S., and Uchiyama, M., (1997), Dynamic
Analysis of Parallel Manipulators under the Singularity-Consistent
Parameterization, Robotica, Vol.~15, pp. 375-384.
\par
Samet, H., (1979), Connected component labeling using quadtrees,
Computer Science Department, University of Maryland, College Park.
\par
Sefrioui, J., and Gosselin, C., (1992), Singularity Analysis and
representation of planar parallel manipulators'' Robotic and
Autonomous Systemes 10, pp. 209-224.
\par
Wenger, Ph., and Chablat, D., (1997), Uniqueness Domains in the
Workspace of Parallel Manipulators, IFAC-SYROCO, Vol. 2, pp
431-436, 3-5 Sept., Nantes.
\par
Wenger, Ph., and El Omri J., (1996), Changing Posture for Cuspidal
Robot Manipulators, International Conference on Robotics and
Automation, Minneapolis, Minnesota, pp. 3173-3178.}
\end{document}